\documentclass[10pt,twocolumn,letterpaper]{article}

\usepackage{iccv}
\usepackage{times}
\usepackage{epsfig}
\usepackage{graphicx}
\usepackage{amsmath}
\usepackage{amssymb}
\usepackage{pifont}
 \usepackage{array,multirow}
 \usepackage{float}
 \usepackage[toc,page]{appendix}

\usepackage[pagebackref=true,breaklinks=true,letterpaper=true,colorlinks,bookmarks=false]{hyperref}

\newcommand{\smallsec}[1]{\vspace{0.2em}\noindent\textbf{#1}}
\newcommand{\xmark}{\ding{55}}%

\iccvfinalcopy 

\ificcvfinal\fi
\begin{document}

\title{Learning to Track with Object Permanence}

\author{\text{\qquad Pavel Tokmakov} \text{\qquad Jie Li} \text{\qquad Wolfram Burgard} \text{\qquad Adrien Gaidon}\\
Toyota Research Institute\\
{\tt\small first.last@tri.global}
}

\maketitle
\ificcvfinal\thispagestyle{empty}\fi

\begin{abstract}
   Tracking by detection, the dominant approach for online multi-object tracking, alternates between localization and association steps. As a result, it strongly depends on the quality of instantaneous observations, often failing when objects are not fully visible. In contrast, tracking in humans is underlined by the notion of object permanence: once an object is recognized, we are aware of its physical existence and can approximately localize it even under full occlusions. 
   In this work, we introduce an end-to-end trainable approach for joint object detection and tracking that is capable of such reasoning. We build on top of the recent CenterTrack architecture, which takes pairs of frames as input, and extend it to videos of arbitrary length. To this end, we augment the model with a spatio-temporal, recurrent memory module, allowing it to reason about object locations and identities in the current frame using all the previous history. It is, however, not obvious how to train such an approach. We study this question on a new, large-scale, synthetic dataset for multi-object tracking, which provides ground truth annotations for invisible objects, and propose several approaches for supervising tracking behind occlusions. 
   Our model, trained jointly on synthetic and real data, outperforms the state of the art on KITTI and MOT17 datasets thanks to its robustness to occlusions.
\end{abstract}

\section{Introduction}

\begin{figure}[t]
\begin{center}
  \includegraphics[width=1\columnwidth]{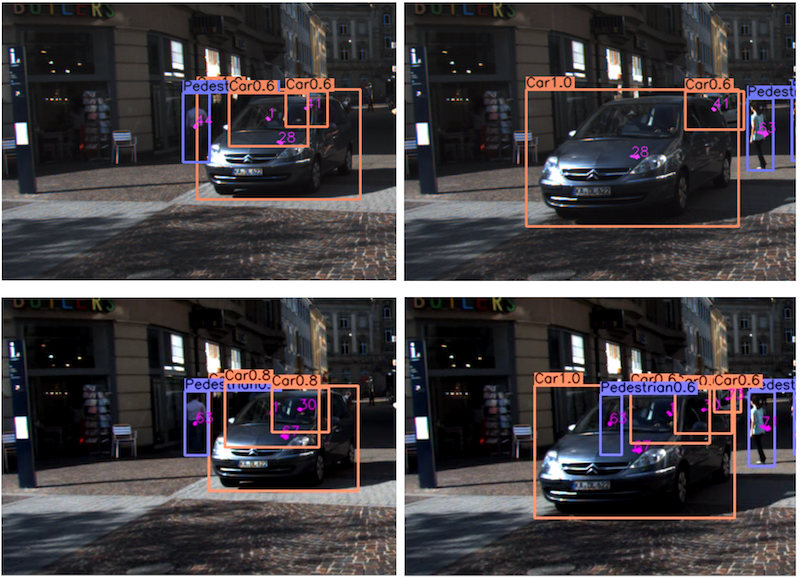}
\end{center}
\vspace*{-2mm}
  \caption{Video frames from the KITTI dataset with outputs of CenterTrack~\cite{zhou2020tracking} (above), and our method (below). By modeling object permanence, our approach is able to hallucinate trajectories of fully occluded instances, such as the person behind the car. }
\vspace*{-2mm}
\label{fig:teaser}
\end{figure}

Consider the video sequence from the KITTI dataset~\cite{geiger2012we} shown in Figure~\ref{fig:teaser}. A man on the left walks behind the moving car and is not visible anymore. Yet, there is no question that he is still there, and did not simply vanish. Moreover, we can approximately infer his location at that moment. This capability is known to cognitive scientists as \textit{object permanence}, and is observed in infants at a very early age~\cite{baillargeon1985object,spelke1990principles}. In adults, understanding that occluded objects do not disappear is important for tasks like driving. In this work, we propose a deep learning-based method for multi-object tracking that is capable of such reasoning.

Virtually all modern multi-object tracking algorithms operate in the tracking-by-detection paradigm. That is, they use an existing object detector to localize objects of interest in every frame of a video, and then link them into tracks, either in an online~\cite{bewley2016simple,breitenstein2009robust}, or in an offline manner~\cite{berclaz2006robust,berclaz2011multiple,braso2020learning,leal2014learning,pirsiavash2011globally}. In this work we focus on the online setting, where a method needs to associate current detections with previously established trajectories~\cite{bewley2016simple,tang2017multiple,wojke2017simple,xu2019spatial}.
A major limitation of these methods is that the localization step is completely independent from the previous history, thus, once an object becomes partially or fully occluded, the detector fails and the trajectory gets broken (see Figure~\ref{fig:teaser}, top).
Recently, several approaches combine detection and tracking in a single model~\cite{bergmann2019tracking,zhou2020tracking}. They take pairs of frames as input and output detections together with pairwise associations. While these methods improve tracking robustness, they can only handle single-frame occlusions.

In this work, we propose an end-to-end trainable, online approach for multi-object tracking that leverages object permanence as an inductive prior. To this end, we first extend the recent CenterTrack architecture~\cite{zhou2020tracking} from pairs of frames as input to \emph{arbitrary video sequences}. The frames are processed by a convolutional gated recurrent unit (ConvGRU)~\cite{ballas2015delving} that encodes the spatio-temporal evolution of objects in the input video, taking the entire history into account. As a result, it can can reason about locations of partially and fully occluded instances using the object permanence assumption (see Figure~\ref{fig:teaser}, bottom).

\begin{figure}[t]
\begin{center}
  \includegraphics[width=0.9\columnwidth]{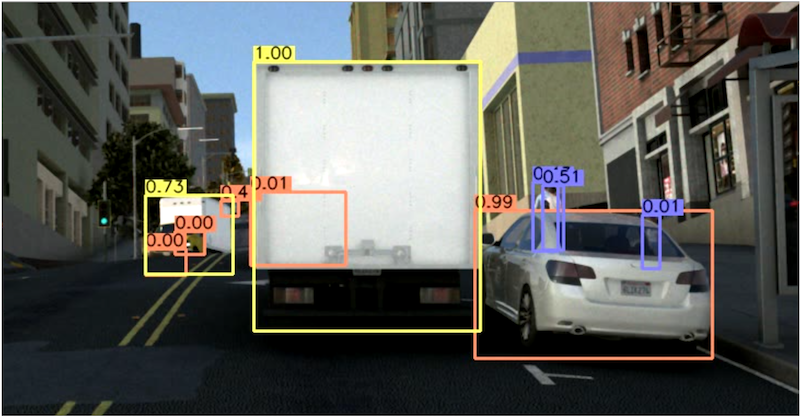}
\end{center}
  \caption{A sample from our synthetic dataset (Section~\ref{sec:datasets}) with ground truth visibility labels. Unlike real datasets, virtual ones provide accurate annotations for all the objects, irrespective of their visibility. }
\label{fig:video}
\vspace{-4mm}
\end{figure}

Supervising  this  behavior  is  a  major  challenge. No tracking dataset currently provides consistent annotations for occluded objects at a large scale (see Figure~\ref{fig:data}) because of the associated uncertainty.
In this paper, we instead propose to use \emph{synthetic data}. Using the Parallel Domain (PD)~\cite{parallel_domain} simulation platform, we generate a dataset of synthetic videos that automatically provides accurate labels for all objects, irrespective of their visibility (see Figure~\ref{fig:video}). We then use this dataset to analyze various approaches for supervising tracking behind occlusions with both ground-truth and pseudo-ground-truth labels for occluded instances. 

Despite the progress in simulation, the domain gap between synthetic and real videos may limit performance. As we show in Section~\ref{sec:abl}, a model directly trained on synthetic videos indeed underperforms when applied to real-world multi-object tracking benchmarks such as KITTI~\cite{geiger2012we}. However, we find that this domain gap can be overcome by just training our model jointly on synthetic and real data, supervising invisible objects only on synthetic sequences. This allows to learn complex behaviors across domains, including trajectory hallucination for fully occluded instances although this was never labeled in the real world.

\smallsec{Our contributions} are three-fold. (1) We propose an end-to-end trainable architecture for joint object detection and tracking that operates on videos of arbitrary length in Sec.~\ref{sec:method}. (2) We demonstrate how this architecture can be trained to hallucinate trajectories of fully invisible objects in Sec.~\ref{sec:train}. (3) We show how to supervise our method with a mix of synthetic and real data in Sec.~\ref{sec:adapt}, and validate it on the KITTI~\cite{geiger2012we} and MOT17~\cite{milan2016mot16} real-world benchmarks in Sec.~\ref{sec:exp}, outperforming the state of the art.
Source code, models, and data are publicly available at \url{https://github.com/TRI-ML/permatrack}. 

\section{Related Work}
Our approach addresses the problem of \textit{multi-object tracking} by designing a joint model for \textit{object detection and tracking in videos} and training it on \textit{synthetic data}. Below, we review the most relevant works in each of these fields.

\smallsec{Multi-object tracking} is the problem of localizing objects from a predefined list of categories in a video and associating them over time based on identity. Most existing approaches treat these two tasks separately, in a paradigm known as tracking-by-detection. The main difference between the methods in this category is whether the association step is performed online or offline.

State-of-the-art online object trackers~\cite{bergmann2019tracking,bewley2016simple,wojke2017simple,xu2019spatial,zhou2020tracking} keep a set of active trajectories as they progress through a video. In every frame, a new list of object detections is processed by either associating them with an existing trajectory, or starting a new one. Early approaches, such as SORT~\cite{bewley2016simple}, used Kalman filter to associate detections based on bounding box overlap, or appearance features from a deep network~\cite{wojke2017simple}. Recent methods proposed to utilize more complex features for association, such as human pose~\cite{tang2017multiple}, or trajectory representations learned with spatio-temporal graph convolutions~\cite{xu2019spatial}. Although some of the methods in this category use a linear motion model~\cite{breitenstein2009robust,mitzel2010multi,yu2007multiple} to propagate a trajectory hypothesis behind occlusions, this heuristic ignores changes in the scene context. In contrast, our approach learns to hallucinate trajectories of occluded objects in an end-to-end manner, outperforming the aforementioned heuristic.

On the other hand, offline approaches~\cite{berclaz2006robust,berclaz2011multiple,braso2020learning,leal2014learning} first build a spatio-temporal graph spanning the whole video, with object detections as nodes~\cite{berclaz2006robust}. Edge costs are defined based on overlap between detections~\cite{jiang2007linear,pirsiavash2011globally,zhang2008global}, their appearance similarity~\cite{braso2020learning,milan2016online,ristani2018features,xu2019spatial}, or motion-based models~\cite{alahi2016social,chen2018real,choi2010multiple,leal2014learning,ren2018collaborative}. The association can then be formulated as maximum flow~\cite{berclaz2011multiple} or, minimum cost problem~\cite{jiang2007linear,leal2014learning}. While these methods can handle complex scenarios, they are not practical due to their non-casual nature and computational complexity. In contrast, our approach does not require future frames and runs in real time.

A few methods have recently attempted to combine object detection and tracking in a single end-to-end learnable framework. Bergman \emph{et al.}~\cite{bergmann2019tracking} utilize the box regression branch in the Faster RCNN detector~\cite{ren2015faster} to propagate objects from frame $t-1$ to $t$. 
Zhou \emph{et al.}~\cite{zhou2020tracking} take a pair of frames as input, and directly output the detections and association vectors, resulting in a simpler architecture. However, both these methods only capture short-term object correspondences. Our approach builds on top of~\cite{zhou2020tracking}, but modifies it to model spatio-temporal evolution of objects in video sequences of arbitrary length, and uses synthetic data to learn to detect and track even under full occlusions. 

Several classical methods~\cite{grabner2010tracking,huang2005tracking,papadourakis2010multiple} have attempted to capture the notion of object permanence through heuristics, such as correlation between the motion of visible and invisible instances in ~\cite{grabner2010tracking}. However, these rule-based methods lack in flexibility. Recently, Shamsian \emph{et al.}~\cite{shamsian2020learning} proposed a fully learning-based approach with promising results on toy, synthetic examples. In contrast, our method is capable of handling full occlusions in the wild.

\smallsec{Video object detection} is primarily concerned with improving the robustness of detectors in videos. Early approaches processed frames individually, but used a Siamese network to establish association between detections and smooth their scores~\cite{feichtenhofer2017detect}. Later, Kang \emph{et al.}~\cite{kang2017object} proposed to pass a stack of several frames to a network and directly output short object tubelets. Finally, Xiao \emph{et al.}~\cite{xiao2018video} augmented an object detector with a spatio-temporal memory module, allowing it to process videos of arbitrary length. Notice that none of these methods tackled the problem of multi-object tracking. Instead, they used short-term associations to improve detection robustness. Similarly to~\cite{xiao2018video}, our architecture also combines an object detector with a spatio-temporal memory module, however, we use a more recent CenterPoint~\cite{zhou2019objects} detector framework and train the model to hallucinate trajectories of fully occluded objects using synthetic data.

\smallsec{Synthetic data} has been used in the past to circumvent the need for manually labeling images~\cite{desouza2019generating,richter2016gtav,ros2016synthia} or videos~\cite{dosovitskiy2015flownet, gaidon2016virtual, tokmakov2019learning}. Most approaches have focused on the setting in which no labels are available in the real world, leveraging unsupervised domain adaptation techniques such as adversarial training~\cite{ganin2015unsupervised,hoffman2018cycada,spigan,vu2019dada} and self-training~\cite{iast,Saporta2020ESLES,crst,cbst}. Although significant progress has been achieved, the performance of these models remains significantly below their counterparts trained on real data in a fully supervised way. In videos, the most successful approaches combined large amounts of synthetic data with small sets of labeled real videos~\cite{ilg2017flownet,tokmakov2019learning}. In this work, we follow a similar route and utilize synthetic data to obtain the expensive labels for occlusion scenarios, while relying on visible object annotations in multi-object tracking datasets to minimize the domain gap. Synthetic datasets with object track labels have been proposed in the past~\cite{fabbri2018learning,gaidon2016virtual}. We choose to capitalize on the recent progress in simulation tools and collect a new, more realistic one.

\section{Methodology}
\subsection{Background}
We build our method on top of the recent CenterTrack~\cite{zhou2020tracking} architecture. Their approach addresses tracking from a local perspective. In particular, CenterTrack takes a pair of frames $\{I^{t-1},I^t\}$ as input together with $H^{t-1}$ - an encoding of locations of previously detected objects in frame ${t-1}$. Objects are represented with their center points $\mathbf{p} \in \mathbb{R}^2$, thus $H^{t-1}$ is compactly encoded with a heatmap. The three input tensors are concatenated and passed through a backbone network $f$, producing a feature map $F^t=f(H^{t-1},I^{t-1},I^t)$, which is used to both localize the object centers in the current frame $\{\hat{\mathbf{p}}_0^t, \hat{\mathbf{p}}_1^t, ...\}$, regress their bounding box sizes $\{\hat{s}_0^t, \hat{s}_1^t, ...\}$, and predict their displacement vectors with respect to the previous frame $\{\hat{\mathbf{d}}_0^t, \hat{\mathbf{d}}_1^t, ...\}$. At test time, displacement vectors are used to project each center to the previous frame via $\hat{\mathbf{p}}_i^t - \hat{\mathbf{d}}_i^t$, and then greedily match it to the closest available center $\hat{\mathbf{p}}_*^{t-1}$, thus recovering the tracks (see~\cite{zhou2020tracking} for more details).

\begin{figure*}[t]
\begin{center}
   \includegraphics[width=1.0\linewidth]{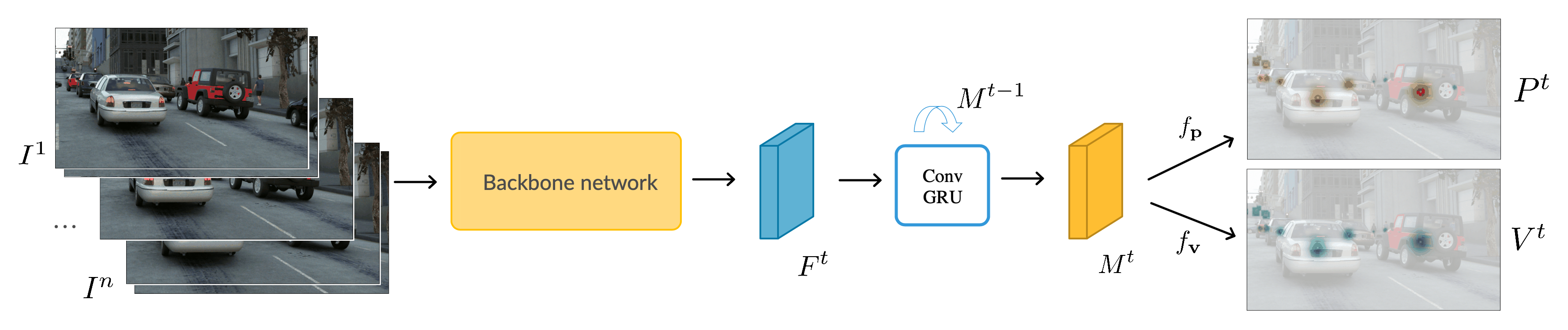}
\end{center}
\vspace{-3mm}
   \caption{Our method takes a sequence of frames as input and processes them individually with a backbone network (shown in yellow). The resulting feature maps (shown in blue) are passed to the ConvGRU module which aggregates a representation of the scene, encoding all the previously seen objects, even if they are fully occluded in the current frame. The memory state at time $t$, (shown in gold), is then used to decode object centers, and estimate their visibility (other outputs directly adapted from~\cite{zhou2020tracking} are not shown for readability).}
\label{fig:method}
\vspace{-3mm}
\end{figure*}

The local nature of CenterTrack is both its strength and its weakness. While only considering a pair of consecutive frames simplifies the architecture of the model, it limits its representational power. In particular, it cannot capture the notion of object permanence in videos, and learn to localize and associate objects under full occlusions. To address this limitation, we first extend~\cite{zhou2020tracking} to a video-level model in Section~\ref{sec:method}. We then describe how to train such a model to track invisible objects using synthetic data in Section~\ref{sec:train}, and detail our domain adaptation approach is Section~\ref{sec:adapt}.

\subsection{A video-level model for tracking}
\label{sec:method}
Our model, shown in Figure~\ref{fig:method}, takes a sequence of frames $\{I^{1},I^2, ..., I^n\}$ as input. Each frame is passed through the backbone $f$ individually to obtain feature maps $\{F^{1},F^2, ..., F^n\}$, which, per CenterTrack formalism, encode the locations of \textit{visible} objects in that frame - an instantaneous representation. To learn a permanent representations of objects in a video, we augment our network with a convolutional gated recurrent unit (ConvGRU)~\cite{ballas2015delving}. It is an extension of the classical GRU~\cite{cho2014learning}, which replaces a 1D state vector with a 2D feature map $M$, and fully connected layers, used to compute state updates, with convolutions.

At each time step $t$, the corresponding feature map $F^t$ is passed to the ConvGRU, together with the previous state $M^{t-1}$ to compute the updated state $M^t = GRU(M^{t-1}, F^t)$. 
Intuitively, the state matrix $M$ represents the entire history of the previously seen objects $\{o_1, o_2, ...\}$ in frames $\{1, ..., t-1\}$ and is updated with the encoding of the visible objects in frame $t$ via a series of learnable, multiplicative transformations (see~\cite{ballas2015delving} for further details). It can thus model the spatio-temporal evolution of objects in the input video sequence by guiding their localization and association in frame $t$ using previous history. Moreover, it can predict locations of the objects that were seen in the past, but are currently occluded. 
Notice that with this architecture there is no need to pass the explicit encoding of the previous frame centers $H^{t-1}$, since they are already captured in the ConvGRU state $M^{t - 1}$.

In practice, to generate the tracks on-line, $M^t$ is recurrently passed through separate sub-networks $f_\mathbf{p}, f_{off}, f_s, f_\mathbf{d}$, which following~\cite{zhou2020tracking}, are used to decode the bounding boxes of the objects and link them into tracks. We augment those with a new visibility head $f_\mathbf{v}$, which produces an output map $V^t \in [0, 1]^{H \times W}$. It is a binary classifier, predicting whether the object center at a particular location corresponds to a visible, or a fully occluded instance. See, for example, the person walking behind the red SUV on the right in Figure~\ref{fig:method}. His location behind occlusion is supervised as a positive for the localization head $P^t$, but as a negative for $V^t$. This distinction is important for evaluation, since annotations for invisible objects are not provided (we also remove them from the validation set of PD for a fair comparison). To avoid being penalized for `false positive' predictions, we only output bounding boxes that are classified as visible by our model, but use the invisible ones to recover object identities after occlusions.

All the operations detailed here are fully differentiable, and thus the model can be trained in an end-to-end fashion with backpropagation through time~\cite{werbos1990backpropagation}. Following~\cite{zhou2020tracking}, we use focal loss~\cite{lin2017focal} to supervise $P^t$ and $V^t$, and L1 loss for the latter three heads. The overall training objective is:
\begin{equation}
    L = \frac{1}{N} \sum_{t=1}^N L^t_\mathbf{p} + L^t_\mathbf{v} + \lambda_{off} L^t_{off}  + \lambda_s L^t_{s} + \lambda_\mathbf{d} L^t_\mathbf{d},
\label{eq:loss}
\end{equation}
where $N$ is the length of the input sequence, and $\lambda_{off}, \lambda_s, \lambda_\mathbf{d}$ are hyper-parameters that balance the contribution of the corresponding losses in the overall objective.

As we discuss in Section~\ref{sec:abl}, training our model  on visible objects alone results in noticeable improvements over~\cite{zhou2020tracking} due to increased robustness to noise in instantaneous observations. Next, we discuss our approach to supervising fully occluded objects.
\begin{figure*}[t]
\begin{center}
  \includegraphics[width=1.0\linewidth]{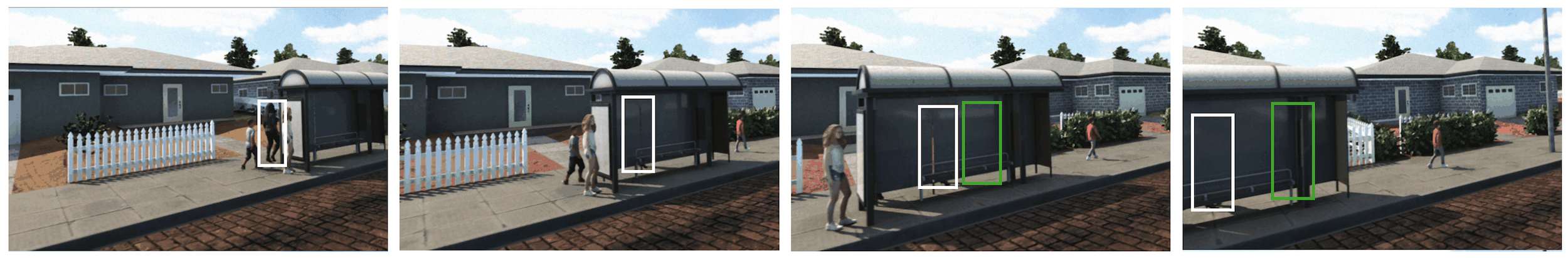}
\end{center}
\vspace{-3mm}
  \caption{Illustration of the ambiguity of ground truth object locations under full occlusions. The woman, shown in white, walks behind the bus stop and then stops. Instead of trying to predict this random event, we propose to supervise the model with deterministic pseud-ground-truth in such scenarios (shown in green, matches the ground truth in the first two frames).}
 \label{fig:occlsup}
 \vspace{-3mm}
\end{figure*}

\subsection{Learning to track behind occlusions}
\label{sec:train}

\subsubsection{Disambiguating levels of visibility}

To generate training labels for a video sequence of length $N$, our method takes object annotations $\{O^{1},O^2, ..., O^N\}$, with $O^t = \{o_1^t, o_2^t, ..., o_m^t\}$ as input. Each object $o_i^t = (\mathbf{p}, \mathbf{s}, id, vis)$ is described by its center $\mathbf{p} \in \mathbb{R}^2$, bounding box size $\mathbf{s}  \in \mathbb{R}^2$, identity $id \in \mathbb{I}$, which is used, together with $\mathbf{p}$, to supervise the displacement vectors $\mathbf{d}$, and visibility level $vis \in [0,1]$, indicating what fraction of the object is visible in the current frame. Naively, one could simply ignore the visibility levels and supervise all the objects in every frame. This would, however, result in the model being forced to hallucinate objects trajectories before they first become visible (e.g. the car driving in front of the truck in Figure~\ref{fig:video} for the whole duration of the video). As we show in our experiments (Table~\ref{tab:invis_anal}), such supervision is effectively label noise and decreases model's performance.

To avoid this, we pre-process the annotations to only start supervising occluded objects after they have been  visible for at least 2 frames. This is the minimal sufficient time for the model to both localize an object and estimate its velocity - the required information for predicting its position under occlusion. 

Concretely, we introduce two thresholds $T_{vis}$ and $T_{occl}$. Then, starting from the first frame in a sequence $O^{1}$, for every object $o_i^1$, if $vis_i^1 < T_{vis}$ the object is treated as a negative, if $T_{vis} < vis_i^1 < T_{occl}$ it is ignored (the model is not penalized for predicting it), and finally, if $vis_i^1 > T_{occl}$ its marked as visible and used to produce the labels. The same procedure is repeated for every frame in a sequence, with the only difference that, starting from frame 3, objects that were marked as visible for two consecutive frames in the past are treated as positives regardless of their visibility status in the current frame. This procedure reduces the ambiguity of supervision for the model. In particular, the second threshold $T_{occl}$ allows for a soft transition between visible and invisible objects.

\subsubsection{Supervising the invisible}

The ambiguity of the location of an invisible object is not fully addressed by the algorithm above. Consider the ground truth trajectory of a person shown in white in Figure~\ref{fig:occlsup}. She walks behind the bus stop, and then stops. In the absence of observations it is impossible for the model to predict this behavior. 
Consequently, such examples also constitute label noise. 
In fact, the only deterministic assumption both a person and a neural network can make about the trajectory of an occluded object is that it will maintain constant velocity. 

Propagating the last observed object location with its constant velocity in the camera frame is also the state of the art approach for handling occlusions in multi-object tracking literature~\cite{bergmann2019tracking,zhou2020tracking}.  It is however, not robust to changes in camera motion. Instead, we propose to generate pseudo-ground-truth labels for supervising our model by propagating the occluded object locations with their last observed velocity in 3D, and projecting the resulting centers to the camera frame, which is made possible by the availability of the full ground truth information in our synthetic dataset. 

Concretely, for an object $i$ getting occluded at time $t$, we take its ground truth centers in the previous two frames in the world coordinate system $\mathbf{P}^{t-1}_i, \mathbf{P}^{t-2}_i \in \mathbb{R}^3$, 
and compute the object velocity $\mathbf{V}_i=\mathbf{P}^{t-1}_i - \mathbf{P}^{t-2}_i$. We then use it to estimate the location of the object center at time $t$ under the constant velocity assumption via $\tilde{\mathbf{P}}^{t}_i = \mathbf{P}^{t-1}_i + \mathbf{V}_i$. Finally, this estimated center is projected to the camera frame via $\tilde{\mathbf{p}}^t_i = \mathbf{K} [\mathbf{R}|\mathbf{t}] \tilde{\mathbf{P}}^{t}_i$, where $\mathbf{K}$ is the camera intrinsics, and $[\mathbf{R}|\mathbf{t}]$ is the camera extrinsic matrix, and is used to replaced the ground-truth center $\mathbf{p}^t_i$ in the corresponding label set $o^t_i$. The same procedure is repeated for all the frames during which the object remains invisible. This principled approach, shown in green in Figure~\ref{fig:occlsup}, results in a deterministic supervision for invisible objects. 

If the object does not re-appear at the expected location, the model is supervised to keep hallucinating the box as described above. Re-identification is then performed at the time of dis-occlusion $t$ by supervising the displacement vector $d_t$ to project the ground-truth, visible location to the hallucinated location at $t-1$.

\subsection{Bridging the sim-to-real domain gap}
\label{sec:adapt}

Analyzing the approaches for supervising tracking behind occlusion described above requires a large video dataset with objects densely labeled regardless of whether they are visible, together with precise visibility scores, 3D coordinates and camera matrices. No real dataset with such labels exists due to the cost and complexity of collecting these annotations.

Instead of going the expensive route, we use synthetic data for which it is easy to automatically generate physically accurate ground truth. However, generalization to real videos remains a challenge.
A few approaches that used synthetic videos for training in the past addressed the domain discrepancy by simply finetuning the resulting model on a small real dataset~\cite{gaidon2016virtual, ilg2017flownet,tokmakov2019learning}. Recall, however, that real datasets in our scenario do not provide consistent annotations for occluded objects, thus such finetuning would result in forgetting the trajectory hallucination behaviour. 

To mitigate this issue, we propose to jointly train our approach on synthetic and real data, where at each iteration a batch is sampled from one of the datasets at random. Moreover, we cannot use real sequences of length more than 2 for the same reason mentioned above (we want the invisible objects supervision to remain consistent). As a result, we sample synthetic clips of length $N$ and real ones of length 2 during training, with the final loss being:
\begin{equation}
    L = \frac{1}{N} \sum_{t=1}^N L^{t}_{sym} + \frac{1}{2} \sum_{t=1}^2 L^{t}_{real},
\label{eq:loss_joint}
\end{equation}
where $L_{sym}$ and $L_{sym}$ are defined in the same way as in Equation~\ref{eq:loss}.
Effectively, we use synthetic videos to learn the desired behavior, and real frame pairs to minimize the domain discrepancy.  

\section{Experiments}
\label{sec:exp}

\subsection{Datasets and evaluation}
\label{sec:datasets}

We use two real datasets in the experimental analysis to compare to prior work: KITTI~\cite{geiger2012we} and MOT17~\cite{milan2016mot16}. In addition, we use a virtual dataset - ParallelDomain (PD), to learn to track behind occlusions. Details of these datasets are provided in the appendix. While our work is focused on the 2D setting, our approach can be directly extended to 3D tracking, as we show on the nuScenes~\cite{caesar2019nuscenes} benchmark in the appendix.

\smallsec{Evaluation metrics} Traditionally, multi-object tracking datasets have been using the CLEAR MOT metrics for evaluation~\cite{bernardin2008evaluating}, with MOTA being the main metric used to compare the methods. It has recently been shown, however, that it overemphasizes detection over association~\cite{luiten2020hota}. 
Instead, supplementary metrics, such as fraction of tracks that are maintained for at least 80\% of their duration (`mostly tracked') have to be used in conjunction with MOTA. 

The lack of a single metric that combines detection and association accuracy has been addressed by track intersection-over-union (IoU) based metrics~\cite{dave2020tao,russakovsky2015imagenet,yang2019video}, and HOTA~\cite{luiten2020hota}. In our analysis we use the former due to its larger emphasis on association accuracy. To formally define track IoU, let $G = \{g_1, \dots, g_T\}$ and $D = \{d_1, \dots, d_T\}$  be a groundtruth and a corresponding predicted track for a video with $T$ frames. Importantly, only one predicted track can be assigned to each ground truth trajectory, and all the unassigned ones are treated as false positives. Track IoU is then defined as:
$\text{IoU}(D, G) = \frac{\sum_{t=1}^T g_t \cap d_t}{\sum_{t=1}^T g_t \cup d_t}$. 
Similarly to standard object detection metrics, track IoU together with track confidence can be used to compute mean average precision (mAP) across categories using a predefined IoU threshold. Following~\cite{dave2020tao} we use a threshold of 0.5, referring to this metric as Track AP. When comparing to the state of the art, we report the standard metrics for each dataset.

\subsection{Implementation details}
For the components of our model shared with CenterTrack~\cite{zhou2020tracking} we follow their architecture and protocol exactly. Here we only provide the values of the new hyper-parameters. Additional details, such as joint training on synthetic and real data, are reported in the appendix.

The ConvGRU has a feature dimension of 256, and uses $7 \times 7$ convolutions. We train the model on synthetic sequences of length 17 for 21 epochs with a batch size 16 using the Adam optimizer. The choice of the clip length is defined by GPU memory constraints. The learning rate is set to $1.25\mathrm{e}{-4}$ and decreased by a factor of 10 every 7 epochs for 1 epoch. It is then increased back to the original value. We have found such a periodic schedule to speed up convergence.  We set the visibility threshold $T_{vis}$ to 0.05 and the occlusion threshold $T_{occl}$ to 0.15, corresponding to 5\% and 15\% of the object being visible respectively. During evaluation, our model is applied in a fully online way, processing all the frames from a video one by one. It runs on a single Tesla V100 GPU at around 10FPS.

\begin{table}[bt]
 \centering
  {
\resizebox{\linewidth}{!}{
    \begin{tabular}{l|c@{\hspace{1em}}c@{\hspace{1em}}c@{\hspace{1em}}|c@{\hspace{1em}}c@{\hspace{1em}}|c@{\hspace{1em}}}
     & GRU & $H^{t-1}$ & Input & Car AP          & Person AP           & mAP              \\\hline
    CenterTrack  & - & \checkmark &  2 fr & 66.2       & 54.4       & 60.3     \\
    \hline
    Ours  & $3\times3$ & \checkmark & 2 fr
        & 64.6       & 49.7       & 57.1       \\
    Ours  & $7\times7$ & \checkmark & 2 fr
        & 65.2       & 54.0       & 59.6        \\
    Ours  & $7\times7$ & \xmark & 2 fr
        & 65.7       & 55.6       & 60.6     \\
   Ours  & $7\times7$ & \xmark & Video
        & \textbf{66.8}       & \textbf{57.9}       & \textbf{62.4}    \\
\end{tabular}
}
}
\caption{Analysis of the architecture of our our model using Track AP on the validation set of PD. We ablate the effect of the filter size in ConvGRU, explicitly passing the encoding of object centers in the previous frame $H^{t-1}$, and training on videos vs frame pairs.}
\vspace{-4mm}
\label{tab:arch_anal}
\end{table}

\subsection{Ablation analysis}
\label{sec:abl}
In this section, we analyze our proposed approach. 
All the variants are trained using exactly the same hyper-parameters and learning rate schedules. Invisible object annotations are ignored in the validation set of PD, so only visible parts of the trajectories are used for evaluation. This allows to fairly compare methods that do and do not have access to invisible object labels during training.

\smallsec{Model variants.} We begin by studying the variants of our video-level tracking model. To this end, we first train CenterTrack on PD using the code provided by the authors and report the results in row 1 of Table~\ref{tab:arch_anal}. The basic variant our our model, shown in row 2, like CenterTrack, takes pairs of frames as input together with $H^{t-1}$, the encoding of detections in the previous frame, but processes them recurrently with a ConvGRU. It performs significantly worse than the baseline, due to the fact that CenterTrack uses a deep network with a large field-of-view to combine the two frames and establish associations between the objects, whereas our model relies on a few convolutional layers in the ConvGRU with $3\times3$ filters. Increasing the filter size (row 3 in Table~\ref{tab:arch_anal}) indeed results in a noticeable performance improvement.
\begin{table}[bt]
 \centering
  {
    
\resizebox{\linewidth}{!}{
    \begin{tabular}{l|l@{\hspace{1em}}c@{\hspace{1em}}|c@{\hspace{1em}}c@{\hspace{1em}}|c@{\hspace{1em}}}
     & Invis. sup. & Post-proc. & Car AP          & Person AP           & mAP              \\\hline
    Ours  & All GT & - 
        & 66.0       & 58.3       & 62.2       \\
    Ours  & Filtered GT & - & 71.1       & 60.6       & 65.9        \\
    Ours  & 2D const v & -
        & 70.7       & 60.8       & 65.7     \\
   Ours  & 3D const v &   -
        & \textbf{71.0}       & \textbf{63.0}       & \textbf{67.0}    \\
    \hline
     CenterTrack  & - & 2D const v &  67.6       & 54.9       & 61.2     \\
    Ours  & 3D const v &   2D const v
        & \textbf{72.7}       & \textbf{63.1}       & \textbf{67.9}    \\
\end{tabular}
}
}
\caption{Comparison of different approaches for handling full occlusions using Track AP on the validation set of PD. We evaluate several supervision strategies, and compare the best variant to the heuristic-based constant velocity track propagation.}
\vspace{-4mm}
\label{tab:invis_anal}
\end{table}

Next, we observe that the additional $H^{t-1}$ input is redundant in our case. Moreover, removing this input allows to avoid the corresponding heat map augmentation strategy proposed by~\cite{zhou2020tracking}. This simplification (show in row 4 in Table~\ref{tab:arch_anal}) further improves the performance of our model, matching CenterTrack. Finally, training and evaluating on longer sequences (last row in Table~\ref{tab:arch_anal}) unlocks the full potential of our architecture to capture object permanence in videos. Even when trained only on visible objects, this variant significantly outperforms the baseline due to its robustness to the noise in instantaneous observations.

\smallsec{Tracking behind occlusions.} In Table~\ref{tab:invis_anal} we now compare various strategies for supervising full occlusion scenarios. Firstly, we observe that the naive approach of training the model to detect and track all the invisible object results in a slight decrease in performance compared to variant trained only on visible ones (last row in Table~\ref{tab:arch_anal}). As discussed in Section~\ref{sec:train}, such supervision is highly ambiguous, as the model cannot localize the fully occluded objects it has not seen before. Accounting for this fact with our proposed annotation filtering strategy (second row in Table~\ref{tab:invis_anal}) results in a 3.5 mAP improvement, validating its importance. 

Next, we compare using ground truth location of invisible object to pseudo-ground-truth obtained via propagating the occluded object with its constant velocity in 2D, or in 3D (see Section~\ref{sec:train} for details). Constant velocity in 2D (row 3 in Table~\ref{tab:invis_anal}) is not robust to camera changes and results in a lower performance than the ground truth locations. In contrast, propagating the target with its last observed velocity in 3D world coordinates (row 4 in the table), results in labels which are both consistent with the observations and fully deterministic, further improving the performance.

Finally, we compare our learning-based approach to the constant velocity post-processing, which is a common way of handling occlusions in the tracking literature~\cite{bergmann2019tracking,zhou2020tracking}, in the last two rows in Table~\ref{tab:invis_anal}.
Firstly, this heuristic-based step does indeed improve the performance of~\cite{zhou2020tracking}, but it remains 5.8 mAP points below our principled method. Secondly, applying this post-processing to the outputs of our method also improves its performance. We do not use any post-processing in the remainder of the experiments.

\begin{figure}
    \centering
    \includegraphics[width=0.8\linewidth]{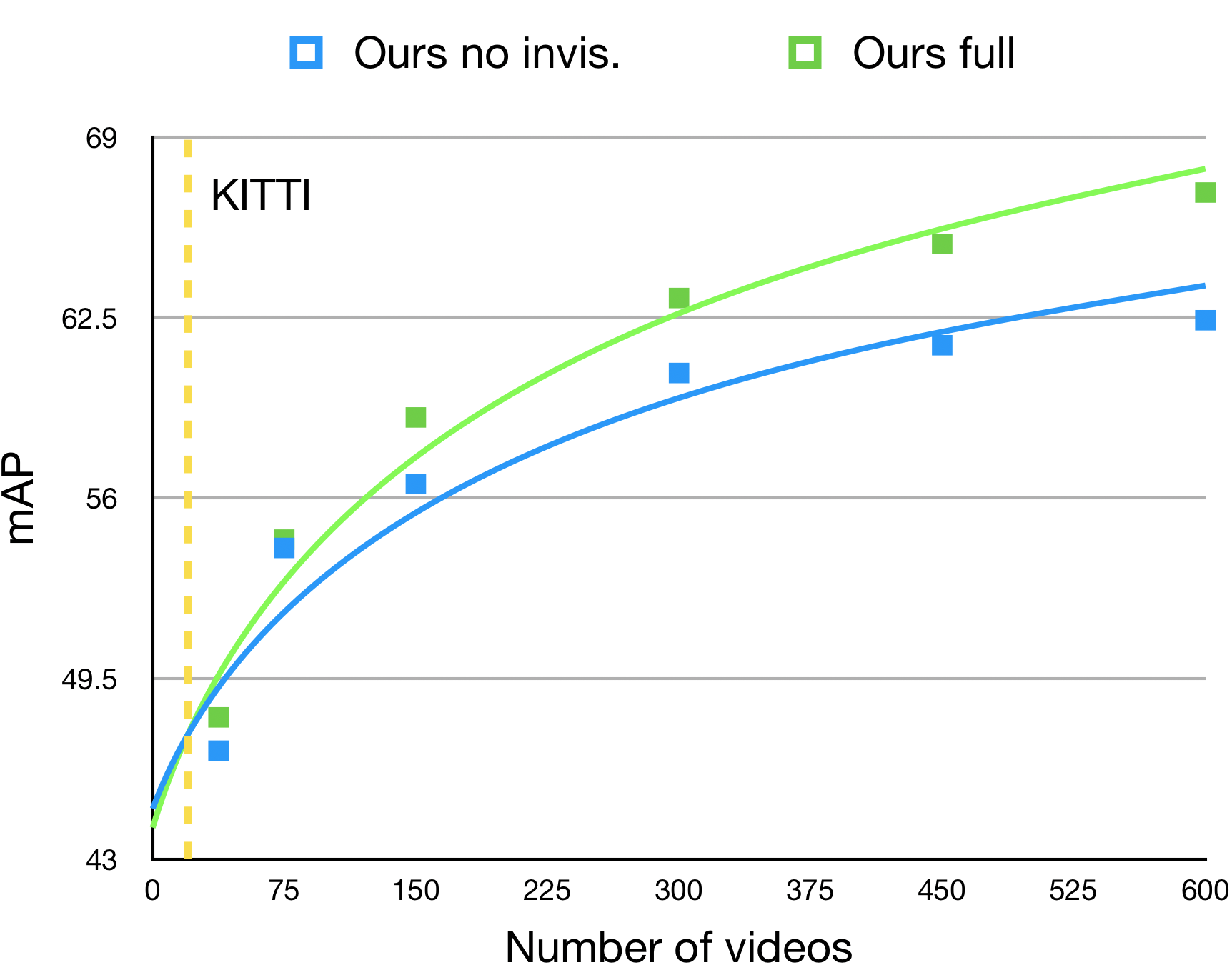}
    \caption{Evaluation of the effect of the number of training videos using Track mAP on the validation set of PD. The gap between our full approach (shown in green), and the variant trained without invisible object labels (shown in blue), increases consistently with the dataset size. This demonstrates that a large dataset is required to learn to hallucinate trajectories of fully occluded objects, and KITTI (shown in yellow) would not be sufficient.}
    \label{fig:data}
    \vspace{-4mm}
\end{figure}

\smallsec{Effect of the dataset size.} In Figure~\ref{fig:data} we plot the validation performance of two variants of our model: the full one, and the one trained on visible objects only, while increasing the number of videos in the training set of PD. One can easily see that the gap between the two variants consistently increases. In fact, below 75 videos it is close to 0. This demonstrates that a large number of examples is required to learn the tracking behind occlusions behaviour.

Yellow dashed line corresponds to the number of videos in the training set of KITTI, and illustrates that even if this dataset provided all required annotations, it would not be sufficient to train our model. MOT17 contains only 7 training videos, making it even less practical for our purposes.

\smallsec{Domain adaptation.} We now demonstrate how the model learned on the large-scale, synthetic PD dataset can be transferred to a real-world benchmark such as KITTI in Table~\ref{tab:domain_anal}, and compare to the CenterTrack model released by the authors. Firstly, we directly evaluate our model trained on PD (fourth row in Tabel~\ref{tab:invis_anal}) and report the results in the second row of Table~\ref{tab:domain_anal}. Despite the significant domain gap between the synthetic and real videos, this variant manages to outperform CenterTrack without seeing a single frame from KITTI. The gap in performance on the Person category remains large, due to its higher visual variability.

Directly fine-tuning our model on KITTI (shown in the third row in the table), helps to reduce the domain discrepancy, improving the Person performance, but also results in un-learning the tracking behind occlusions behaviour, as reflected in the drop in Car AP.  In contrast, jointly training on the two datasets achieves the best results overall, validating the effectiveness of the proposed approach.

Finally, we apply our domain adaptation approach to the CenterTrack model trained on PD (first row in Table~\ref{tab:arch_anal}), and report the results in the last row of Table~\ref{tab:domain_anal}. One can see that synthetic data also improves CenterTrack results, but they remain 6.1 mAP point below those of our approach. This demonstrates that the improvements mainly come from our model's ability to better handle occlusions.
\begin{figure*}
\begin{center}
  \includegraphics[width=1.0\linewidth]{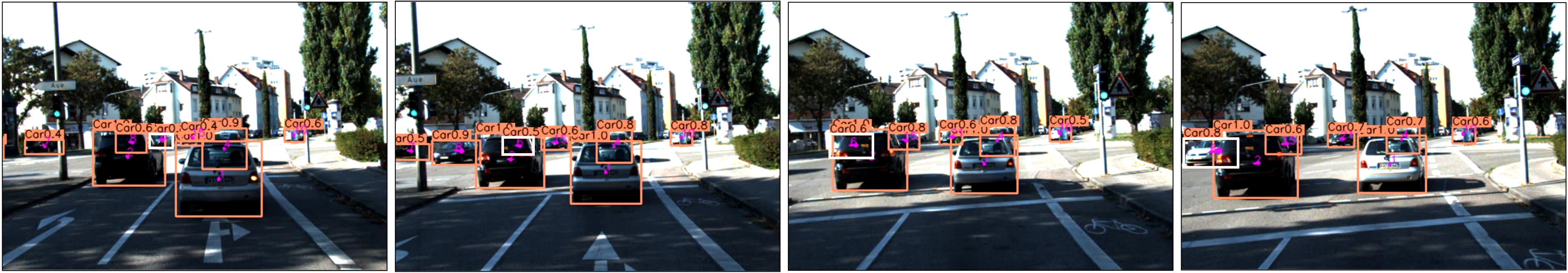}
\end{center}
 \vspace{-2mm}
  \caption{Qualitative results on a test sequence from the KITTI benchmark. Our approach is able to successfully track a fully occluded car on the left ({\tt id 6}, highlighted in white). Please see supplementary video for a more comprehensive analysis on both datasets.}
 \label{fig:qual}
 \vspace{-2mm}
\end{figure*}

\begin{table}[bt]
 \centering
  {
   
\resizebox{\linewidth}{!}{
    \begin{tabular}{l|l@{\hspace{1em}}c@{\hspace{1em}}|c@{\hspace{1em}}c@{\hspace{1em}}|c@{\hspace{1em}}}
     & KITTI & PD & Car AP          & Person AP           & mAP              \\\hline
    CenterTrack~\cite{zhou2020tracking}  & \checkmark & \xmark  & 73.8       & 39.4       & 56.6     \\
     \hline
    Ours  & \xmark & \checkmark
        & 83.3       & 38.2       & 60.8       \\
    Ours (tune)  & \checkmark & \checkmark
        & 75.7       & 44.6       & 60.2       \\
    Ours (joint)  & \checkmark & \checkmark & \textbf{84.7}       & \textbf{56.3}       & \textbf{70.5}        \\
    CenterTrack (joint)  & \checkmark  & \checkmark
        & 77.2       & 51.6       & 64.4     \\
\end{tabular}
}
}
\caption{Domain adaptation analysis using Track AP on the validation set of KITTI. We demonstrate the effectiveness of our simple strategy, and confirm that the improvements mainly come from better occlusion handling.}
\vspace{-4mm}
\label{tab:domain_anal}
\end{table}

\subsection{Comparison to the state of the art}
In this section, we compare our full method, which we refer to as PermaTrack, to the state of the art on the KITTI~\cite{geiger2012we}, and MOT17~\cite{milan2016mot16} benchmarks.  All our models are pre-trained on PD and adapted to the real domain using the training set of the corresponding dataset.
\begin{table*}[bt]
  \begin{center}
    \begin{tabular}{l|c@{\hspace{1em}}c@{\hspace{1em}}c@{\hspace{1em}}c@{\hspace{1em}}c@{\hspace{1em}}|c@{\hspace{1em}}c@{\hspace{1em}}c@{\hspace{1em}}c@{\hspace{1em}}c@{\hspace{1em}}}
     & \multicolumn{5}{c|}{Car} &
          \multicolumn{5}{c}{Person}\\
    & HOTA $\uparrow$  & MOTA $\uparrow$         & MT $\uparrow$ & PT $\downarrow$  & ML$\downarrow$           & HOTA $\uparrow$ & MOTA $\uparrow$         & MT $\uparrow$  & PT $\downarrow$  & ML $\downarrow$       \\\hline
     MASS~\cite{karunasekera2019multiple}    & 68.3 &   84.6 &  74.0 & 23.1 & 2.9  & - & - & -  & - & -    \\
    IMMDP~\cite{Xiang2015ICCV}   & 68.7 &   82.8  & 60.3 & 27.5 &	12.2  & - &	- &	- & - & -     \\
    AB3D~\cite{weng2019baseline}  & 69.8  &    83.5  &   67.1   & 21.5 &	11.4 &  35.6  &    38.9 &	17.2 &	41.6 &	41.2 \\
    TuSimple~\cite{choi2015near}   & 71.6  & 86.3  & 71.1 & 22.0 &	6.9 & 45.9  & 57.6 & 30.6 & 44.3 &	25.1   \\
    SMAT~\cite{gonzalez2020smat}   & 71.9  &  83.6  & 62.8 & 31.2 & 6.0 &  - &	- &	-   & - & -  \\
    TrackMPNN~\cite{rangesh2021trackmpnn}   & 72.3  & 87.3  & 84.5 & 13.4 &	\bf{2.2}    & 39.4 & 52.1 &	35.1 & 46.1 &	18.9   \\
    CenterTrack~\cite{zhou2020tracking}   & 73.0  & 88.8  &   82.2 &  15.4 & 2.5   &  40.4  & 53.8 & 35.4 &  43.3 & 21.3      \\
    \hline
    PermaTrack (Ours) & \bf{78.0}  & \bf{91.3}  &  \bf{85.7} &  \bf{11.7} & 2.6 & \bf{48.6}  & \bf{66.0} & \bf{48.8} &  \bf{35.4} & \bf{15.8} \\

\end{tabular}

\caption{Comparison to the state of the art on the test set of the KITTI benchmark using aggregate metrics. Some methods only report results on Car. Our approach outperforms all the other methods by a significant margin on all the metrics except for ML on Car.}
\vspace{-4mm}
\label{tab:kitti_test}
 \end{center}
\end{table*}

\smallsec{KITTI.} Table~\ref{tab:kitti_test} lists the results on the KITTI test set, comparing to vision-based, online methods. Our method outperforms the state of the art on all metrics on both categories, except for ML (Mostly Lost) on Car, where we are 0.4 points below~\cite{rangesh2021trackmpnn}. Notice that we are 5.7 point above this method on the main HOTA metric, and are outperforming it by a large margin on all metrics on Person. Our improvements over the state-of-the-art CenterTrack are 5 HOTA points on Car and 8.4 on Person, which is notable, given that typical differences between the methods on the leader-board are within 1 point, and that CenterTrack is pre-trained on a large-scale, \textit{real-world} nuScenes dataset~\cite{caesar2019nuscenes}.

\begin{table}[bt]
  \begin{center}
  {
      \normalsize
\resizebox{\linewidth}{!}{
    \begin{tabular}{c|l|c|c@{\hspace{1em}}c@{\hspace{1em}}c@{\hspace{1em}}c@{\hspace{1em}}c@{\hspace{1em}}}
    & & T.R. &   IDF1 $\uparrow$ & MOTA $\uparrow$          & MT  $\uparrow$  & PT  $\downarrow$          & ML  $\downarrow$          \\\hline
     \parbox[t]{2mm}{\multirow{4}{*}{\rotatebox[origin=c]{90}{Public}}} & CenterTrack~\cite{zhou2020tracking} & \xmark &  63.2 & 63.1  &   37.5 & 38.1    & 24.5   \\
    & PermaTrack (Ours) & \xmark  & 67.0 & 67.8  &  \bf{43.7}  &  \bf{36.3}  & 20.1  \\
      & CenterTrack~\cite{zhou2020tracking} & \checkmark  & 66.4 & 63.8 &   37.2 & 38.1    & 24.8   \\
    & PermaTrack (Ours) & \checkmark  & \bf{71.1}  &  \bf{68.2} &  41.0  &  39.5  & \bf{19.5}  \\
    \hline
 \parbox[t]{2mm}{\multirow{4}{*}{\rotatebox[origin=c]{90}{Private}}} &   CenterTrack~\cite{zhou2020tracking} & \xmark  & 64.2 & 66.1   &   41.3 & 37.5    & 21.2   \\
    & PermaTrack (Ours)  & \xmark   & 68.2 & 69.4   & \bf{46.3}  &  \bf{36.0}  & \bf{17.7}  \\
     & CenterTrack~\cite{zhou2020tracking} & \checkmark  & 69.4  & 66.2   &   39.5 & 38.3    & 22.1   \\
    & PermaTrack (Ours)  & \checkmark  & \bf{71.9} & \bf{69.5}   &  42.5  &  \bf{39.8}  & \bf{17.7}  \\
\end{tabular}
}
}

\caption{Comparison to the state of the art on the validation set of the MOT17 using private and public detections. Our method outperforms the state-of-the-art CenterTrack approach on all metrics both with and without the Track Rebirth (T.R.) post-processing.}
\vspace{-4mm}
\label{tab:mot_test}
 \end{center}
\end{table}

\smallsec{MOT17.} The policy on this dataset is that only methods that don't use external data for training can be evaluated on the test server with public detections. With just 7 videos in the training set, MOT17 is not sufficient to learn the complex tracking behind occlusions behaviour. For fairness, we compare our method to the variant of the state-of-the-art CenterTrack which is pre-trained on the \textit{real-world} CrowdHuman dataset~\cite{shao2018crowdhuman} using the validation set of MOT17 in Table~\ref{tab:mot_test}. The results on the test set using private detections are reported in the appendix.

Without post-processing, our approach outperforms~\cite{zhou2020tracking} using both public and private detections. The improvements are especially significant in the public evaluation (the same bounding boxes used by both methods), emphasizing our method's better tracking abilities. Finally, adding Track Rebirth post-processing from~\cite{zhou2020tracking} (T.R. in the table, a variant of constant velocity) improves the performance of both approaches and does not change the conclusions. 

\smallsec{Qualitative results.} We provide an example of the output of our method on a sequence from the test set of KITTI in Figure~\ref{fig:qual}. In this challenging scenario, as the ego-vehicle drives forward, the car on the left ({\tt id 6}, highlighted in white) is fully occluded by another moving vehicle ({\tt id 2}), but our method manages to correctly localize it and maintain the trajectory. A more comprehensive analysis of our approach is presented in the supplementary video\footnote{\url{https://www.youtube.com/watch?v=Dj2gSJ_xILY}}.

\section{Conclusion}
In this work, we propose PermaTrack, an end-to-end-trainable approach for joint object detection and tracking. Thanks to its recurrent memory module, it is capable of reasoning about the location of objects using the entire previous history, and not just the current observation. Combined with supervision from synthetic data, this allows to train the model to track objects when they are fully occluded - a key concept in cognitive science known as object permanence. 

Our method obtains state-of-the-art results on the KITTI and MOT17 multi-object tracking benchmarks. 
While the ablation analysis demonstrates that hallucinating trajectories of invisible objects is a crucial factor in the final performance, knowledge about the full history also increases robustness to partial occlusions, and other low visibility scenarios, such as motion blur.

{\small
\bibliographystyle{ieee_fullname}
\bibliography{egbib}
}

\clearpage
\appendix
\appendixpage

In this appendix, we provide additional visualizations, experimental results and implementation details that were not included in the main paper due to space limitations. We begin by describing the contents of the supplementary video, which includes qualitative examples of our algorithm's output in Section~\ref{sec:vd}. In Section~\ref{sec:data} we provide details about the datasets used in our work, and further elaborate on the PD dataset in Section~\ref{sec:pd}. A real world nuScenes~\cite{caesar2019nuscenes} dataset for 3D tracking could potentially be used to train our method directly, and we demonstrate preliminary results on it in Section~\ref{sec:nu}. We conclude by reporting all the metrics on KITTI~\cite{geiger2012we} and MOT17~\cite{milan2016mot16} benchmarks in Section~\ref{sec:mtr} and listing the remaining implementation details in Section~\ref{sec:impl}.

\section{Qualitative analysis}
\label{sec:vd}

We demonstrate the outputs of our method on several videos from KITTI and MOT17 datasets\footnote{\url{https://www.youtube.com/watch?v=Dj2gSJ_xILY}}. We show raw outputs of the model, \textit{without} any post-processing steps, such as constant velocity propagation. 

\smallsec{00:08-00:20} In the video version of the example from Figure 5 in the main paper it is easier to see how our approach is able to correctly localize the moving car ({\tt id 6}) occluded by the black vehicle at an intersection. Despite the complexity of the scenario (both cars, as well as the ego vehicle, are in motion) our approach successfully tracks this object as it undergoes a full occlusion.

\smallsec{00:24-00:49} In this example, the grey car ({\tt id 5}) arrives at an intersection and is repeatedly occluded by three other vehicles. Again, our approach is able to maintain its trajectory throughout the whole sequence of occlusions.

\smallsec{00:53-01:09} The final example from the KITTI test set demonstrates occlusions by people. First, the person with {\tt id 18} is occluded by a group of people, but his trajectory is not broken. Then the group continues forward to hide the car with {\tt id 6}, and the person with  {\tt id 18} is re-occluded by another pedestrian, but both cases are successfully handled by our method.

\smallsec{01:13-01:35} In this sequence from the validation set of MOT17 it is worth paying attention to the group of people in the back on the left. They get occluded by another group, but this complex, multi-target occlusion scenario is also effectively solved by our learning-based approach.

\smallsec{01:39-01:50} The final positive example shows a scenario which differs from our synthetic dataset: a camera with a top-down view is flying over a street as a group of people is getting occluded by a pole. Nevertheless, our method is able to generalize to this challenging sequence. 

\smallsec{01:54-02:05} Here we demonstrate a failure mode of our method: the car with {\tt id 5} is occluded by a grey wagon and, for a moment, their centers overlap. CenterPoint~\cite{zhou2019objects} detector architecture, which serves as a basis of our model, can only predict one object center at every location. As a result, the wagon is not detected in this frame, and an id switch happens between the two trajectories. Such mistakes can often be fixed by a short-term constant velocity post-processing.

\smallsec{02:09-02:17} Another failure mode is shown in this example. In the complex, indoor scene shot from a person's perspective the agent with {\tt id 45} occludes two people on the left. The furthest of them is tracked for a few frames, but is eventually lost. This is due to the fact that the person was partially occluded in the beginning, so the initial confidence of the model was low, illustrating a limitation of our approach.

\section{Datasets}
\label{sec:data}
\smallsec{KITTI} is a multi-object tracking benchmark capturing city driving scenarios~\cite{geiger2012we}. It consists of 21 training and 29 test sequences. Cars, pedestrians, and cyclists are annotated with 2D bounding boxes at 10 FPS. Following prior work, we evaluate on the former two categories. For ablation analysis, we split each training sequence in half, and use the first half for training and the second for validation. The test set is reserved for comparison to the state of the art.

\smallsec{MOT17} is the standard benchmark for people tracking~\cite{milan2016mot16}. Unlike KITTI, most of the videos are captured with a static camera, and feature crowded indoor and outdoor areas. It consists of 7 training and 7 test sequences annotated with 2D bounding boxes at 25-30 FPS.  As for KITTI, we split the training videos in half to create a validation set. The standard policy on this dataset is to only report methods that do not use external data on the test set with public detections. For fairness, we compare to the state of the art on the validation set, but also report results with private detections on the test set below.

\smallsec{ParallelDomain (PD)} is our synthetic dataset used for learning to track behind occlusions. 
It was collected using a state-of-the-art synthetic data generation service~\cite{parallel_domain}. The dataset contains 210 photo-realistic videos with driving scenarios in city environments captured at $20$ FPS. Representing crowded streets, these videos feature lots of occlusion and dis-occlusion scenarios involving people and vehicles, providing all aforementioned annotations required by our method. Each video is 10 seconds long and comes with 3 independent camera views, effectively increasing the dataset size to 630 videos. We use 582 of those for training, and the remaining 48 for validation. We ignore invisible object labels in the validation set, and evaluate all the models on visible parts of the trajectories only.

\section{Statistics for the Parallel Domain Dataset}
\label{sec:pd}
Our synthetic dataset is generated through a state-of-the-art synthetic data generation service powered by \textit{Parallel Domain}~\cite{parallel_domain}. The dataset contains $210$ short snippet of crowded urban driving scenarios. Each video is $10$ seconds long and is captured at $20$FPS, providing $3$ independent camera views. We treat the different camera views as independent videos, resulting in $630$ videos in total. We split the videos into a training set with $582$ videos and a validation set with $48$ videos. There are no overlapping scenes between the two sets.

Each frame of a video is annotated with amodal bounding boxes defined for 9 object classes, though we focus on Pedestrians and Cars in this work.
Consistent instance ids are provided across the video to support tracking association.
Both visible and occluded bounding boxes are labeled with a precise visibility scores, indicating the fraction of the object which is visible in the current frame.
The dataset features environments with a variable number of agents, time of day, and weather conditions. 
We summarize the per-class statistic regarding the length of tracks in Table~\ref{tab:pd_stat}. One can see that most trajectories span close to a half of the video, which is crucial for learning long-term tracking behaviour.

\begin{figure}
    \centering
    \includegraphics[width=\linewidth]{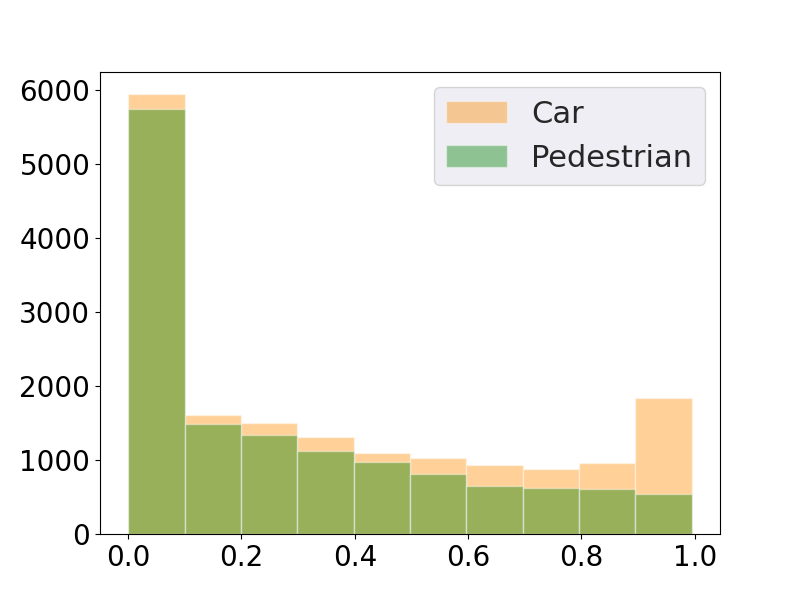}
    \caption{Histogram of occlusion ratios for the tracks in PD. We plot the fraction of the trajectory length during which the visibility score of the boxes is lower than $0.05$ for both \textbf{Pedestrian} and \textbf{Car} categories. $0$ indicates that the object is at least partially visible for the whole duration of the track, while $1$ indicates that it is occluded for the whole duration.}
    \label{fig:pd}
    \vspace{-2mm}
\end{figure}

We additionally provide histograms depicting the distribution of the fraction of trajectories that are occluded within a video for the classes used in our work in Figure~\ref{fig:pd}. This figure demonstrates that $64.9\%$ of Pedestrian and $58.1\%$ of Car trajectories are fully occluded for at least $10\%$ of their duration, providing enough training examples for learning to track with object permanence. 
%
%

The photo-realistic synthetic data along with the amodal annotations in the Parallel Domain dataset allow us to investigate a wide variety of model variants and supervision approaches with accurate annotations. As indicated in our experimental analysis, the dataset can not only be used for prototyping and analysing the proposed algorithm, but also to pre-train models with object permanence awareness that can be successfully transferred to real world datasets through our simple sim-to-real adaptation approach. 

\begin{table}[]
\centering
\resizebox{0.48\textwidth}{!}{%
\begin{tabular}{|c|c|c|c|c|c|c|}
\hline
\multirow{2}{*}{Class} & \multicolumn{2}{c|}{\# of Tracks} & \multicolumn{2}{c|}{Avg. Length} & \multicolumn{2}{c|}{Max Length} \\ \cline{2-7} 
 & Train & Val & Train & Val & Train & Val \\ \hline
Pedestrian & 13056 & 846 & 83.9 & 65.3 & 200 & 200 \\ \hline
Car & 15604 & 1517 & 105.9 & 94.2 & 200 & 200 \\ \hline
Bicyclist & 283 & 12 & 92.0 & 44.4 & 200 & 108 \\ \hline
Bus & 274 & 13 & 118.3 & 78.8 & 200 & 200 \\ \hline
Caravan/RV & 90 & 3 & 112.6 & 71.3 & 200 & 86 \\ \hline
OtherMovable & 1537 & 134 & 107.1 & 83.3 & 200 & 200 \\ \hline
Motorcycle & 223 & 24 & 90.1 & 79.1 & 200 & 192 \\ \hline
Motorcyclist & 246 & 28 & 90.2 & 75.0 & 200 & 200 \\ \hline
Truck & 839 & 76 & 111.2 & 91.4 & 200 & 200 \\ \hline
\end{tabular}%
}
\caption{Parallel Domain per-category dataset statistics. Note that we count the same instances observed from different cameras separately as we treat them independently during training and evaluation.}
\label{tab:pd_stat}
\vspace{-4mm}
\end{table}

\section{Evaluation on nuScenes}
\label{sec:nu}
In this section, we validate our method on the large-scale nuScenes benchmark for 3D tracking. Since this dataset is comparable in scale to PD, and provides full 3D information, we train on it directly, using the approach described in Section 3.3.2 of the main paper (see Section~\ref{sec:impl} for details). In Table~\ref{tab:nu_val}, we compare the performance of our proposed method to the CenterTrack~\cite{zhou2020tracking} baseline on the validation set. Our approach indeed improves the performance on the main metrics. In particular, we improve the AMOTA by 4.1 points, which is a 60\% relative improvement.

Note that our work focuses on 2D tracking and we only report these results to validate that our method can in principle be generalized to the 3D scenario. Tracking objects in 3D is an important problem in itself, and comes with many caveats. For instance, nuScenes is annotated at a very low frame rate (2FPS) and features a significant amount of label noise. Thoroughly investigating the effect of these factors on our method's performance is out of scope of this work.

\begin{table}[bt]
 \centering
  {
    \footnotesize
\resizebox{\linewidth}{!}{
    \begin{tabular}{l|c@{\hspace{1em}}c@{\hspace{1em}}c@{\hspace{1em}}c@{\hspace{1em}}c@{\hspace{1em}}}
       & AMOTA          &MOTA  & Recall                \\\hline
    CenterTrack~\cite{zhou2020tracking}    & 6.8      &6.1&0.23       \\
    PermaTrack (Ours)  & \textbf{10.9}     &\textbf{8.1}&\textbf{0.23}       \\
\end{tabular}
}
}
\caption{Validating that our method can be generalized to 3D object tracking using the validation set of nuScenes. Our method indeed outperforms the CenterTrack baseline by a significant margin on the main metrics, but a thorough investigation of 3D tracking is out of scope of this work.}
\vspace{-2mm}
\label{tab:nu_val}
\end{table}

\section{Full Tables for KITTI and MOT17}
\label{sec:mtr}
In this section we report the final results of our method on KITTI and MOT17 using all the metrics on these benchmarks for reference.

KITTI uses 3 main sets of tracking metrics: HOTA-based metrics~\cite{luiten2020hota}, CLEAR MOT metrics~\cite{bernardin2008evaluating}, and  MT/PT/ML metrics~\cite{li2009learning}. We report them in Tables~\ref{tab:kitti_hota}, \ref{tab:kitti_mota}, and~\ref{tab:kitti_mt} respectively. Full results are available on the challenge website~\cite{kitti_test}.
\begin{table*}[bt]
  \begin{center}
    \begin{tabular}{l|c@{\hspace{1em}}c@{\hspace{1em}}c@{\hspace{1em}}c@{\hspace{1em}}c@{\hspace{1em}}c@{\hspace{1em}}c@{\hspace{1em}}c@{\hspace{1em}}}
    Category & HOTA $\uparrow$  & DetA $\uparrow$         & AssA $\uparrow$ & DetRe $\uparrow$   & DetPr $\uparrow$  & AssRe $\uparrow$ & AssPr $\uparrow$ & LocA $\uparrow$  \\\hline
    Car & 78.0  & 78.3  & 78.4 &  81.7 & 86.5 & 81.1 & 89.5 & 87.1 \\
    Person & 48.6  & 52.3 & 45.6 &  57.4 & 71.0 & 49.6 & 73.3 & 78.6 \\

\end{tabular}
\caption{Results of our method on the test set of the KITTI benchmark using the HOTA metrics.}
\vspace{-4mm}
\label{tab:kitti_hota}
 \end{center}
\end{table*}

MOT17 uses a combination of CLEAR MOT and MT/PT/ML metrics. We report all the metrics on the validation set in Table~\ref{tab:mot_mota}, and on the test set with private detections in Table~\ref{tab:mot_mota_test}. Full results are available on the challenge website~\cite{mot_test}.

\begin{table*}
  \begin{center}
    \begin{tabular}{l|c@{\hspace{1em}}c@{\hspace{1em}}c@{\hspace{1em}}c@{\hspace{1em}}c@{\hspace{1em}}c@{\hspace{1em}}}
    Category & MOTA $\uparrow$  & MOTP $\uparrow$         & MODA $\uparrow$ & IDSW $\downarrow$   & sMOTA $\uparrow$   \\\hline
    Car & 91.3  & 85.7  & 92.1 &  258 & 78.0 & \\
    Person & 66.0  & 74.5 & 67.7 &  403 & 47.1 \\

\end{tabular}
\caption{Results of our method on the test set of the KITTI benchmark using the CLEAR MOT metrics.}
\vspace{-4mm}
\label{tab:kitti_mota}
 \end{center}
\end{table*}

\begin{table}
  \begin{center}
    \begin{tabular}{l|c@{\hspace{1em}}c@{\hspace{1em}}c@{\hspace{1em}}c@{\hspace{1em}}c@{\hspace{1em}}}
    Category & MT $\uparrow$  & PT $\downarrow$         & ML $\downarrow$ & FRAG $\downarrow$    \\\hline
    Car & 85.7  & 11.7 & 2.6 &  250  \\
    Person & 48.8  & 35.4 & 15.8 &  646  \\

\end{tabular}
\caption{Results of our method on the test set of the KITTI benchmark using the MT/PT/ML metrics.}
\vspace{-4mm}
\label{tab:kitti_mt}
 \end{center}
\end{table}

\def\arraystretch{1.5}
\begin{table*}[bt]
  \begin{center}
    \begin{tabular}{c|l|c|c@{\hspace{1em}}c@{\hspace{1em}}c@{\hspace{1em}}c@{\hspace{1em}}c@{\hspace{1em}}c@{\hspace{1em}}c@{\hspace{1em}}c@{\hspace{1em}}c@{\hspace{1em}}c@{\hspace{1em}}}
    & & T.R. &   IDF1 $\uparrow$ & IDP $\uparrow$ & IDR $\uparrow$ & MOTA $\uparrow$  & MOTP $\uparrow$          & MT  $\uparrow$  & PT  $\downarrow$          & ML  $\downarrow$  & IDSW  $\downarrow$   & FRAG  $\downarrow$          \\\hline
     \parbox[t]{2mm}{\multirow{2}{*}{\rotatebox[origin=c]{90}{Public}}}
    & PermaTrack & \xmark  & 67.0 & 77.5 & 59.0 & 67.8 & 0.178  & 43.7  & 36.3 & 20.1 & 0.8\% & 1.0\%  \\
    & PermaTrack & \checkmark  & 71.1 & 82.1 & 62.6 &   68.2 & 0.181 &  41.0  &  39.5  & 19.5 & 0.5\% & 1.1\%  \\
    \hline
 \parbox[t]{2mm}{\multirow{2}{*}{\rotatebox[origin=c]{90}{Private}}} 
    & PermaTrack  & \xmark   & 68.2 & 75.9 & 61.9 & 69.4 & 0.18  & 46.3  &  36.0  & 17.7 & 0.9\% & 1.1\%  \\
    & PermaTrack  & \checkmark  & 71.9 & 81.0 & 64.7 & 69.5 & 0.181 &  42.5  &  39.8  & 17.7  & 0.5\% & 1.1\% \\
\end{tabular}

\caption{Results of our method on the validation set of the MOT17 using private and public detections. T.R. stand for Track Rebirth post-processing from~\cite{zhou2020tracking}.}
\vspace{-4mm}
\label{tab:mot_mota}
 \end{center}
\end{table*}

\begin{table*}[bt]
  \begin{center}
    \begin{tabular}{l|c|c@{\hspace{1em}}c@{\hspace{1em}}c@{\hspace{1em}}c@{\hspace{1em}}c@{\hspace{1em}}c@{\hspace{1em}}c@{\hspace{1em}}c@{\hspace{1em}}}
     & T.R. &   IDF1 $\uparrow$ & IDP $\uparrow$ & IDR $\uparrow$ & MOTA $\uparrow$            & MT  $\uparrow$         & ML  $\downarrow$  & IDSW  $\downarrow$   & FRAG  $\downarrow$          \\\hline
    PermaTrack  & \checkmark  & 68.9 & 75.1 & 63.6 & 73.8  &  43.8   & 17.2  & 3699 & 6132 \\
\end{tabular}

\caption{Results of our method on the test set of the MOT17 using private detections. This variant uses Track Rebirth (T.R.). A subset of metrics is shown, which is reported on the MOT leader-board.}
\vspace{-4mm}
\label{tab:mot_mota_test}
 \end{center}
\end{table*}

\section{Further Implementation Details}
\label{sec:impl}
Learning to localize objects that are not visible in the current frame is challenging, and the model tends to ignore them. To avoid this, we increase the weight of the localization loss by a factor of 20 for fully occluded instances and sample sequences which contain occlusion scenarios with a probability which is proportional to the occlusion length.

For domain adaptation to KITTI~\cite{geiger2012we} and MOT17~\cite{milan2016mot16}, we first pre-train the model on PD, and then fine-tune it jointly on PD and the corresponding dataset using the loss in Equation 2 in the main paper. The batches are sampled from each dataset with an equal probability. We use batch size 16 for all datasets, and train for 5 epochs with a learning rate of $1.25\mathrm{e}{-4}$ using the Adam optimizer. The learning rate is decreased by a factor of 10 after the 4th epoch. An epoch is defined as 5000 iterations for KITTI + PD training, and as 1600 iterations for MOT + PD due to the difference in dataset sizes. 

We have found that, since videos in MOT17 are mostly captured with static cameras and the occlusions are mostly short-term, constant velocity in 2D serves as a reasonable approximation for ground truth locations of occluded people. Based on this observation, we use MOT17 sequences of length 13 during joint fine-tuning, and supervise person locations under occlusions using the pseudo-groundtruth obtained via trajectory interpolation. This strategy simplifies domain adaptation, however, as we have discussed in the main manuscript, the training set of MOT17 is too small to learn the parameters of our model from scratch.

When training on the large scale nuScenes~\cite{caesar2019nuscenes} dataset we do not use PD, and instead generate pseudo-ground truth labels using the approach described in Section 3.3.2 in the main paper. For supervising 3D losses we follow all the details in~\cite{zhou2020tracking} exactly. We have found that due to a significant amount of label noise in nuScenes using a large batch size is crucial for achieving top results. Following~\cite{zhou2020tracking}, we first pre-train our model using sequences of length 2 and batch size 64 for 70 epochs. The learning rate is set to $1.25\mathrm{e}{-4}$ and decreased by  factor of 10 after 60 epochs. We then fine-tune this model using sequences of length 6 and batch size 32 for 10 epochs, decreasing the initial learning rate of $1.25\mathrm{e}{-4}$ by a factor of 10 after the 8th epoch. Finally, we freeze the backbone and further fine-tune this model with sequences of length 17 and batch size 32 to capture longer-term occlusions. This last fine-tuning stage uses the same learning rate schedule as the previous one.

\end{document}